\documentclass[10pt,twocolumn,letterpaper]{article}

\usepackage{cvpr}
\usepackage{times}
\usepackage{epsfig}
\usepackage{graphicx}
\usepackage{subfigure}
\usepackage{amsmath}

\usepackage{amssymb}
\usepackage{color}
\usepackage{multirow}

\usepackage[breaklinks=true,colorlinks,bookmarks=false]{hyperref}

\cvprfinalcopy 

\def\cvprPaperID{1822} 

\definecolor{orange}{rgb}{1,0.5,0}
\definecolor{mdgreen}{rgb}{0.05,0.6,0.05}
\definecolor{mdblue}{rgb}{0,0,0.7}
\definecolor{dkblue}{rgb}{0,0,0.5}
\definecolor{dkgray}{rgb}{0.3,0.3,0.3}
\definecolor{slate}{rgb}{0.25,0.25,0.4}
\definecolor{gray}{rgb}{0.5,0.5,0.5}
\definecolor{ltgray}{rgb}{0.7,0.7,0.7}
\definecolor{purple}{rgb}{0.7,0,1.0}
\definecolor{lavender}{rgb}{0.65,0.55,1.0}

\begin{document}

\title{AlignedReID: Surpassing Human-Level Performance in Person Re-Identification}

\author{	 Xuan Zhang\textsuperscript{1}\footnotemark[1] ,
        		Hao Luo\textsuperscript{1,2}\footnotemark[1] \footnotemark[2] ,
		Xing Fan\textsuperscript{1,2}\footnotemark[2] ,
		Weilai Xiang\textsuperscript{1}, 
        		Yixiao Sun\textsuperscript{1},
		Qiqi Xiao\textsuperscript{1}, \\
		Wei Jiang\textsuperscript{2}, 
	    	Chi Zhang\textsuperscript{1},
		Jian Sun\textsuperscript{1}\\
\textsuperscript{1}Megvii, Inc. (Face++)\\
\textsuperscript{2}Institute of Cyber-Systems and Control, Zhejiang University\\
{  \tt\small\{zhangxuan,xiangweilai,sunyixiao,xqq,zhangchi,sunjian\}@megvii.com } \\
{ \tt\small\{haoluocsc,xfanplus,jiangwei$\_$zju\}@zju.edu.cn  } 
}

\maketitle

\renewcommand{\thefootnote}{\fnsymbol{footnote}}
\footnotetext[1]{Equal contribution}
\footnotetext[2]{The work was done when Hao and Xing were interns at MegVii, Inc. (Face++)}

\begin{abstract}
In this paper, we propose a novel method called AlignedReID that extracts a global feature which is jointly learned with local features.
Global feature learning benefits greatly from local feature learning, which performs an alignment/matching by calculating the shortest path between two sets of local features, without requiring extra supervision. After the joint learning, we only keep the global feature to compute the similarities between images. Our method achieves rank-1 accuracy of $94.4\%$ on Market1501 and $97.8\%$ on CUHK03, outperforming state-of-the-art methods by a large margin. We also evaluate human-level performance and demonstrate that our method is the first to surpass human-level performance on Market1501 and CUHK03, two widely used Person ReID datasets.
\end{abstract}

\section{Introduction}
\label{introduction}
Person re-identification (ReID), identifying a person of interest at other time or place, is a challenging task in computer vision.
Its applications range from tracking people across cameras to searching for them in a large gallery, from grouping photos in a photo album to visitor analysis in a retail store. Like many visual recognition problems, variations in pose, viewpoints illumination, and occlusion make this problem non-trivial.

Traditional approaches have focused on low-level features such as colors, shapes, and local descriptors \cite{farenzena2010person, hamdoun2008person}.
With the renaissance of deep learning, the convolutional neural network (CNN) has dominated this field \cite{matsukawa2016person, varior2016gated, cheng2016person, zheng2016person, lin2017improving, matsukawa2016person}, by learning features in an end-to-end fashion through various metric learning losses such as contrastive loss \cite{varior2016gated}, triplet loss \cite{liu2017end}, improved triplet loss \cite{cheng2016person}, quadruplet loss \cite{chen2017beyond}, and triplet hard loss \cite{hermans2017defense}.

\begin{figure}[t]
\centering
\includegraphics[width=1.0\linewidth]{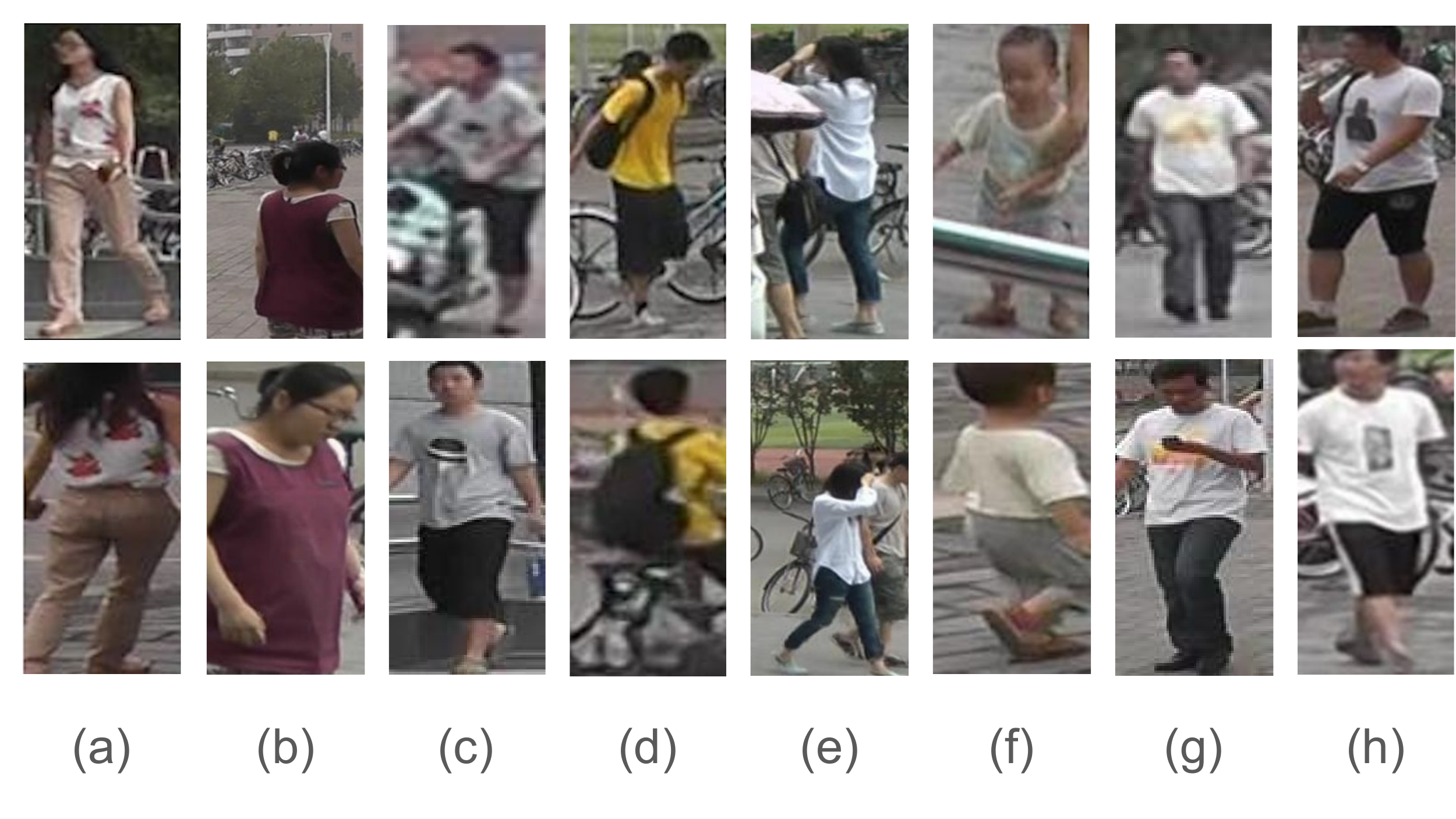}
\caption{Challenges in ReID: (a-b) inaccurate detection, (c-d) pose misalignments,
(e-f) occlusions, (g-h) very similar appearance.}
\label{fig:demo}
\end{figure}

Many CNN-based approaches learn a global feature, without considering the spatial structure of the person. This has a few major drawbacks:
1) inaccurate person detection boxes might impact feature learning, e.g., Figure \ref{fig:demo} (a-b);
2) the pose change or non-rigid body deformation makes the metric learning difficult, e.g., Figure \ref{fig:demo} (c-d); 
3) occluded parts of the human body might introduce irrelevant context into the learned feature, e.g., Figure \ref{fig:demo} (e-f);
4) it is non-trivial to emphasis local differences in a global feature, especially when we have to distinguish two people with very similar appearances, e.g., Figure \ref{fig:demo} (g-h).
To explicitly overcome these drawbacks, recent studies have paid attention to part-based, local feature learning.
Some works \cite{varior2016siamese, xiao2016cross, yao2017deep} divide the whole body into a few fixed parts, without considering the alignment between parts.
However, it still suffers from inaccurate detection box, pose variation, and occlusion.
Other works use pose estimation result for the alignment \cite{zheng2017pose, wei2017glad, zhao2017spindle}, which requires additional supervision and a pose estimation step (which is often error-prone). 

In this paper, we propose a new approach, called AlignedReID, which still learns a global feature, but performs an automatic part alignment during the learning, without requiring extra supervision or explicit pose estimation.
In the learning stage, we have two branches for learning a global feature and local features jointly. In the local branch, we align local parts by introducing a shortest path loss.
In the inference stage, we discard the local branch and only extract the global feature.
We find that only applying the global feature is almost as good as combining global and local features.
In other words, the global feature itself, with the aid of local features learning, can greatly address the drawbacks we mentioned above, in our new joint learning framework.
In addition, the form of global feature keeps our approach attractive for the deployment of a large ReID system, without costly local features matching.    

We also adopt a mutual learning approach \cite{zhang2017deep} in the metric learning setting, to allow two models to learn better representations from each other.
Combining AlignedReID and mutual learning, our system outperforms state-of-the-art systems on Market1501, CUHK03, and CUHK-SYSU by a large margin.
To understand how well human perform in the ReID task, we measure the best human performance of ten professional annotators on Market1501 and CUHK03.
We find that our system with re-ranking \cite{zhong2017re} has a higher level of accuracy than the human.
To the best of our knowledge, this is the first report in which machine performance exceeds human performance on the ReID task.



\begin{figure*}[htb]
\centering
\includegraphics[width=.99\linewidth]{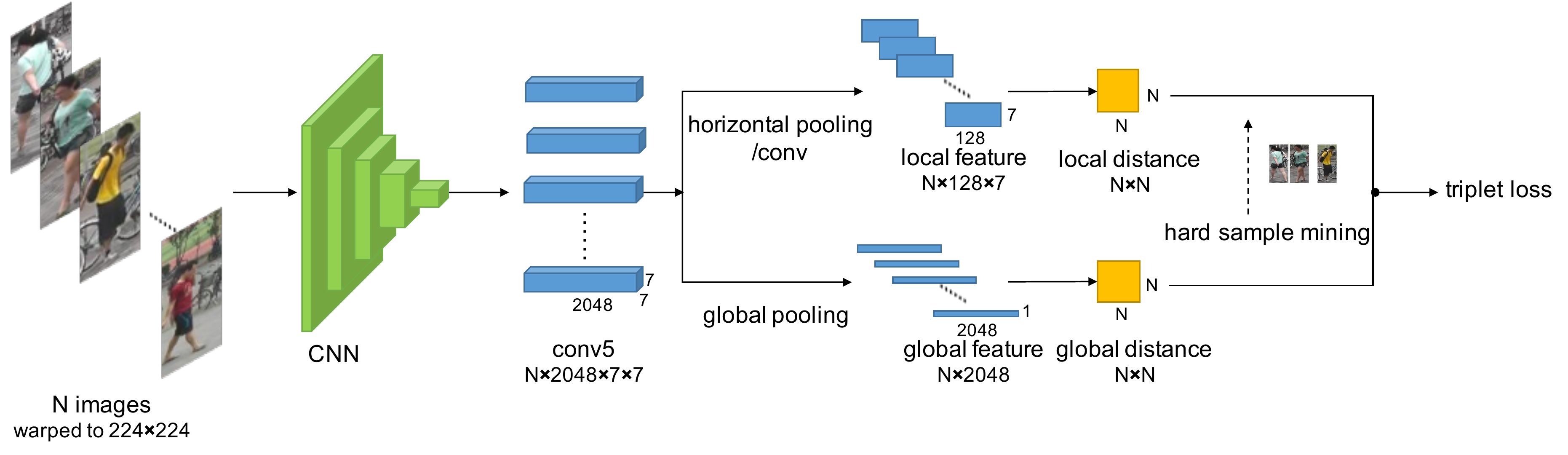}
\caption{The framework of AlignedReID.
Both the global branch and the local branch share the same convolution network to extract the feature map.
The global feature is extracted by applying global pooling directly on the feature map.
For the local branch, one $1\times1$ convolution layer is applied after horizontal pooling, which is a global pooling with a horizontal orientation.
Triplet hard loss is applied, which selects triplet samples by hard sample mining according to global distances.}
\label{fig:Structure}
\end{figure*}

\section{Related Work}
\label{relatedwork}

\noindent \textbf{Metric Learning}.
Deep metric learning methods transform raw images into embedding features, then compute the feature distances as their similarities. Usually, two images of the same person are defined as a positive pair, whereas two images of different persons are a negative pair.
Triplet loss \cite{liu2017end} is motivated by the margin enforced between positive and negative pairs.
Selecting suitable samples for the training model through hard mining has been shown to be effective \cite{hermans2017defense, chen2017beyond, xiao2017margin}.
Combining softmax loss with metric learning loss to speed up the convergence is also a popular method \cite{geng2016deep}. 

\noindent \textbf{Feature Alignments.}
Many works learn a global feature to represent an image of a person, ignoring the spatial local information of images.
Some works consider local information by dividing images into several parts without an alignment \cite{varior2016siamese, xiao2016cross, yao2017deep},
but these methods suffer from inaccurate detection boxes, occlusion and pose misalignment.

Recently, aligning local features by pose estimation has become a popular approach.
For instance, pose invariant embedding (PIE) aligns pedestrians to a standard pose to reduce the impact of pose \cite{zheng2017pose} variation.
A Global-Local-Alignment Descriptor (GLAD) \cite{wei2017glad} does not directly align pedestrians, but rather detects key pose points and extracts local features from corresponding regions.
SpindleNet \cite{zhao2017spindle} uses a region proposed network (RPN) to generate several body regions, gradually combining the response maps from adjacent body regions at different stages. These methods require extra pose annotation and have to deal with the errors introduced by pose estimation.

\noindent \textbf{Mutual Learning.}
\cite{zhang2017deep} presents a deep mutual learning strategy where an ensemble of students learn collaboratively and teach each other throughout the training process.
DarkRank \cite{chen2017darkrank} introduces a new type of knowledge-cross sample similarity for model compression and acceleration, achieving state-of-the-art performance.
These methods use mutual learning in classification.
In this work, we study mutual learning in the metric learning setting.

\noindent \textbf{Re-Ranking.}
After obtaining the image features, most current works choose the L2 Euclidean distance to compute a similarity score for a ranking or retrieval task. \cite{wang2017deep, zhong2017re, bai2017scalable} perform an additional re-ranking to improve ReID accuracy.
In particular, \cite{zhong2017re} proposes a re-ranking method with $k$-reciprocal encoding, which combines the original distance and Jaccard distance.

\section{Our Approach}
\label{method}
In this section, we present our AlignedReID framework, as shown in Figure \ref{fig:demo}.

\begin{figure}[tb]
\centering
\includegraphics[width=1.05\linewidth]{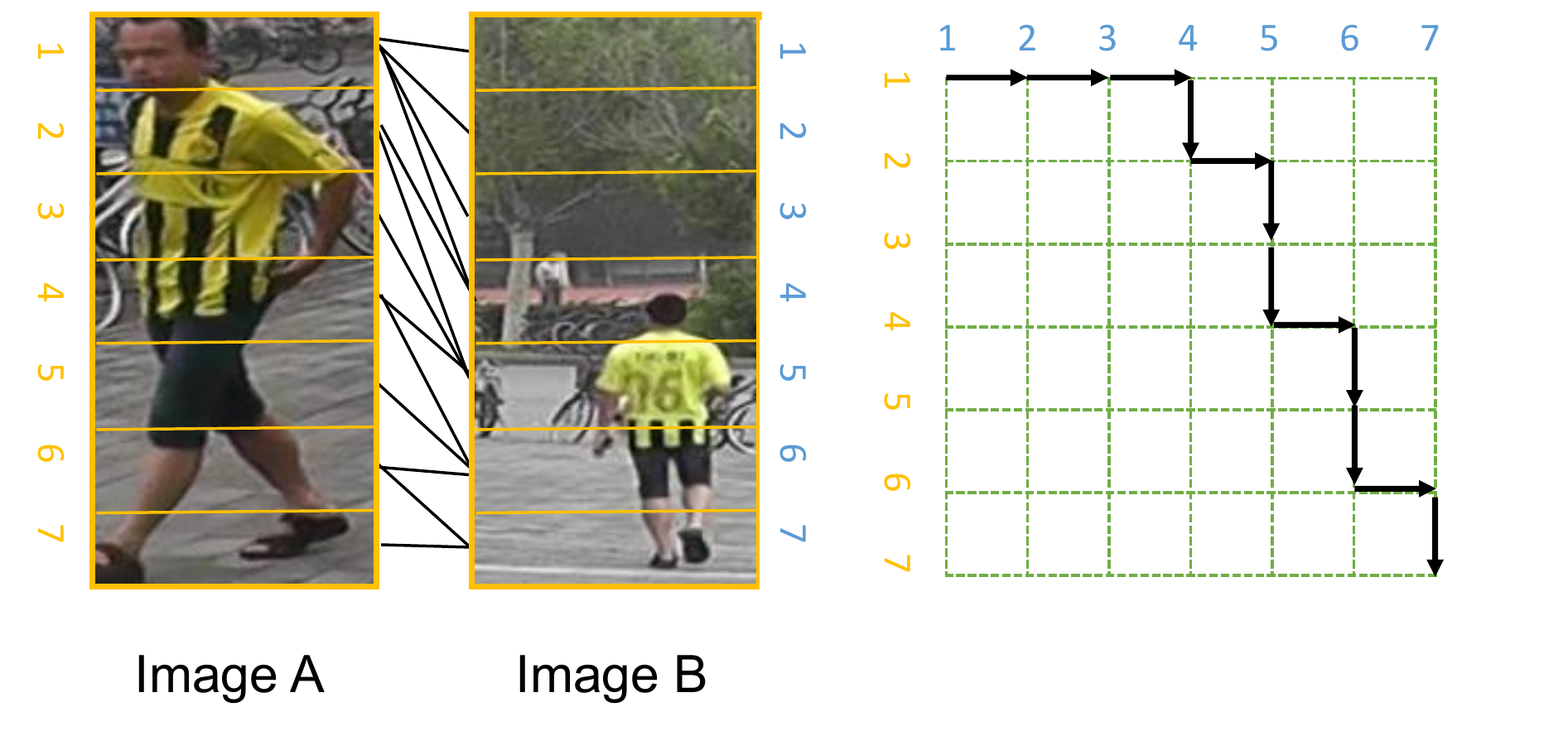}
\caption{Example of AlignedReID local distance computed by finding the shortest path.
The black arrows show the shortest path in the corresponding distance matrix on the right.
The black lines show the corresponding alignment between the two images on the left.}
\label{fig:ShortestPath}
\end{figure}

\begin{figure*}[t]
\centering
\includegraphics[width=.85\linewidth]{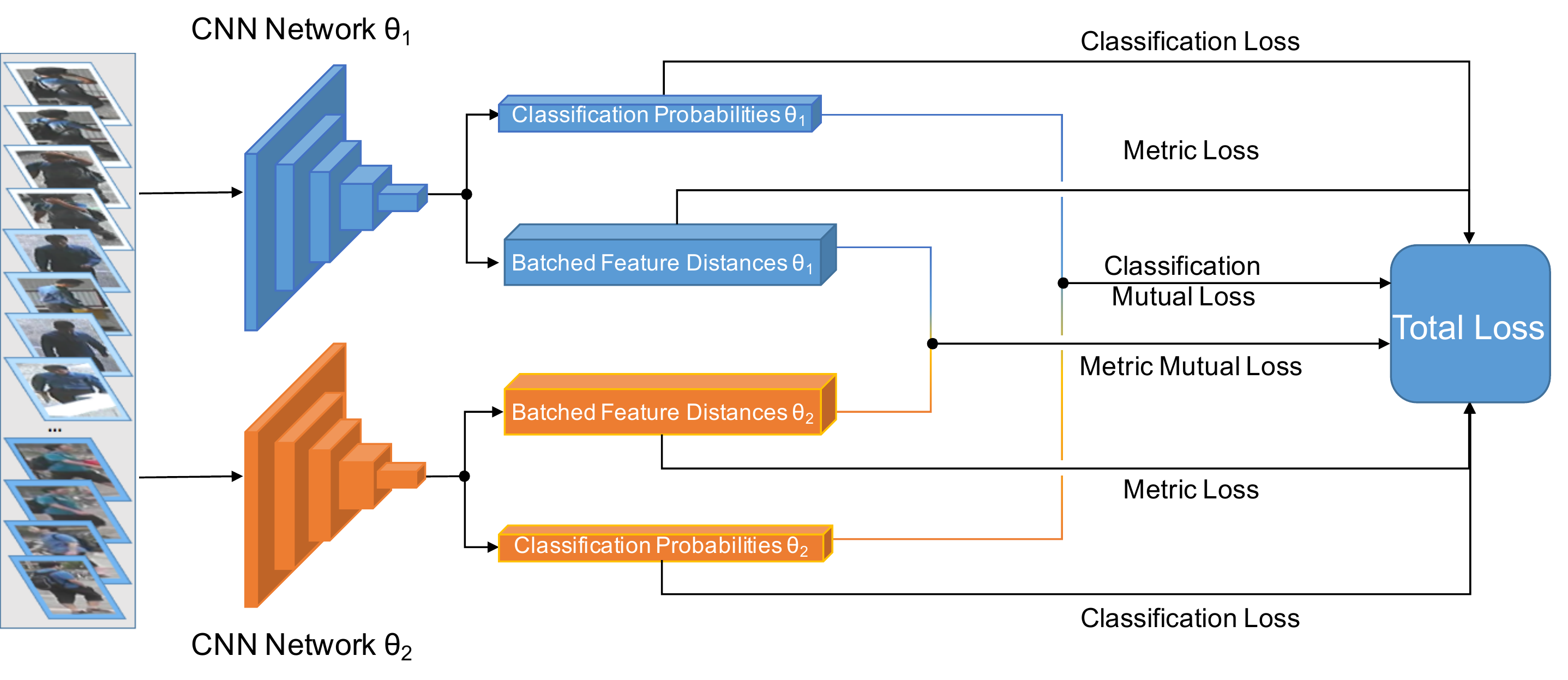}
\caption{Framework of the mutual learning approach. Two networks with parameters $\theta_1$ and $\theta_2$ are trained together. Each network has two branches: a classification branch and a metric learning branch. The classification branches are trained with classification losses, and learn each other through classification mutual loss. The metric learning branches are trained with metric losses, which include both global distance and local distance. Meanwhile, the metric learning branches learn each other by metric mutual loss.}
\label{fig:mutual}
\end{figure*}

\subsection{AlignedReID}
In AlignedReID, we generate a single global feature as the final output of the input image, and use the L2 distance as the similarity metric. However, the global feature is learned \emph{jointly} with local features in the learning stage. 

For each image, we use a CNN, such as Resnet50 \cite{he2016deep}, to extract a feature map, which is the output of the last convolution layer ($C\times H\times W$, where $C$ is the channel number and $H\times W$ is the spatial size, e.g., $2048\times 7\times 7$ in Figure \ref{fig:demo}).
A global feature (a $C$-d vector) is extracted by directly applying global pooling on the feature map.
For the local features, a horizontal pooling, which is a global pooling in the horizontal direction, is first applied to extract a local feature for each row, and a $1\times 1$ convolution is then applied to reduce the channel number from $C$ to $c$.
In this way, each local feature (a $c$-d vector) represents a horizontal part of the image for a person.
As a result, a person image is represented by a global feature and $H$ local features.

The distance of two person images is the summation of their global and local distances.
The global distance is simply the L2 distance of the global features.
For the local distance, we dynamically match the local parts from top to bottom to find the alignment of local features with the minimum total distance.
This is based on a simple assumption that, for two images of the same person, the local feature from one body part of the first image is more similar to the semantically corresponding body part of the other image.

Given the local features of two images, $F = \{f_1, \cdots, f_H\}$ and $G = \{g_1, \cdots, g_H\}$, we first normalize the distance to $[0,1)$ by an element-wise transformation:
\begin{equation}
d_{i,j} = \frac{e^{||f_i-g_j||_2}-1}{e^{||f_i-g_j||_2}+1} \quad i,j \in 1,2,3...,H,
\label{PM}
\end{equation} 
where $d_{i,j}$ is the distance between the $i$-th vertical part of the first image and the $j$-th vertical part of the second image. A distance matrix $D$ is formed based on these distances, where its $(i,j)$-element is $d_{i,j}$. We define the local distance between the two images as the total distance of the shortest path from $(1,1)$ to $(H,H)$ in the matrix $D$. The distance can be calculated through dynamic programming as follows:
\begin{equation}\label{ShortestPath}
S_{i,j}=\begin{cases}
d_{i,j} & i=1,  j=1 \\
S_{i-1,j}+d_{i,j} & i \ne 1,j=1 \\
S_{i,j-1}+d_{i,j} & i=1,j \ne 1 \\
min(S_{i-1,j},S_{i,j-1})+d_{i,j} & i \ne 1, j \ne 1,
\end{cases}
\end{equation}
where $S_{i,j}$ is the total distance of the shortest path when walking from $(1,1)$ to $(i,j)$ in the distance matrix $D$, and $S_{H,H}$ is the total distance of the final shortest path (i.e., the local distance) between the two images.

As shown in Fig. \ref{fig:ShortestPath}, images A and B are samples of the same person.
The alignment between the corresponding body parts, such as part $1$ in image A, and part $4$ in image B, are included in the shortest path.
Meanwhile, there are alignments between non-corresponding parts, such as part $1$ in image A, and part $1$ in image B, still included in the shortest path.
These non-corresponding alignments are necessary to maintain the order of vertical alignment, as well as make the corresponding alignments possible.
The non-corresponding alignment has a large L2 distance, and its gradient is close to zero in Eq.\ref{PM}.
Hence, the contribution of such alignments in the shortest path is small. 
The total distance of the shortest path, i.e., the local distance between two images, is mostly determined by the corresponding alignments.

The global and local distance together define the similarity between two images in the learning stage, and we chose TriHard loss proposed by \cite{hermans2017defense} as the metric learning loss.
For each sample, according to the global distances, the most dissimilar one with the same identity and the most similar one with a different identity is chosen, to obtain a triplet.
For the triplet, the loss is computed based on both the global distance and the local distance with different margins.
The reason for using the global distance to mine hard samples is due to two considerations.
First, the calculation of the global distance is much faster than that of the local distance.
Second, we observe that there is no significant difference in mining hard samples using both distances.

Note that in the inference stage, we only use the global features to compute the similarity of two person images. We make this choice mainly because we unexpectedly observed that the global feature itself is also almost as good as the combined features. This somehow counter-intuitive phenomenon might be caused by two factors: 1) the feature map jointly learned is better than learning the global feature only, because we have exploited the structure prior of the person image in the learning stage; 2) with the aid of local feature matching, the global feature can pay more attention to the body of the person, rather than over fitting the background.  

\subsection{Mutual Learning for Metric Learning}
We apply mutual learning to train models for AlignedReID, which can further improve performance.
A distillation-based model usually transfers knowledge from a pre-trained large teacher network to a smaller student network, such as \cite{chen2017darkrank}.
In this paper, we train a set of student models simultaneously, transferring knowledge between each other, such as \cite{zhang2017deep}.
Differing from \cite{zhang2017deep}, which only adopts the Kullback-Leibler (KL) distance between classification probabilities, we propose a new mutual learning loss for metric learning.

The framework of our mutual learning approach is shown in Fig. \ref{fig:mutual}.
The overall loss function includes the metric loss, the metric mutual loss, the classification loss and the classification mutual loss.
The metric loss is decided by both the global distances and the local distances, while
the metric mutual loss is decided only by the global distances.
The classification mutual loss is the KL divergence for classification as in \cite{zhang2017deep}.

Given a batch of N images, each network extracts their global features and calculates the global distance between each other as an $N\times N$ batch distance matrix,
where $M^{\theta_1}_{ij}$ and $M^{\theta_2}_{ij}$ denote the $(i,j)$-th element in the matrices separately.
The mutual learning loss is defined as
\begin{equation}\label{bddl}
\begin{aligned} 
L_{M} = \frac{1}{N^2}\sum_i^N \sum_j^N & \Big( [ZG(M^{\theta_1}_{ij})-M^{\theta_2}_{ij}]^2 \\
 & + [M^{\theta_1}_{ij}-ZG(M^{\theta_2}_{ij})]^2 \Big),
 \end{aligned}
\end{equation}
where $ZG(\cdot)$ represents the zero gradient function, which treats the variable as constant when calculating gradients, stopping the backpropagation in the learning stage.

By applying the zero gradient function, the second-order gradients is
\begin{equation}\label{LM22}
\frac{\partial^2 L_{M} }{\partial M^{\theta_1}_{ij}\partial M^{\theta_2}_{ij}} = 0.
\end{equation}
We found that it speeds up the convergence and improves the accuracy compared to a mutual loss without the zero gradient function.


\section{Experiments}
\label{experiments}
In this section, we present our results on three most widely used ReID datasets: Market1501 \cite{zheng2015scalable}, CUHK03 \cite{Li2014DeepReID}, and CUHK-SYSU \cite{xiao2016end}.
\subsection{Datasets}

\textbf{Market1501} contains 32,668 images of 1,501 labeled persons of six camera views.
There are 751 identities in the training set and 750 identities in the testing set.
In the original study on this proposed dataset, the author also uses mAP as the evaluation criteria to test the algorithms.

\textbf{CUHK03} contains 13,164 images of 1,360 identities.
It provides bounding boxes detected from deformable part models (DPMs) and manual labeling.


\textbf{CUHK-SYSU} is a large-scale benchmark for a person search, containing 18,184 images (99,809 bounding boxes) and 8,432 identities.
The training set contains 11,206 images of 5,532 query persons, whereas the test set contains 6,978 images of 2,900 persons.

Note that we only train a single model using training samples from all three datasets, as in \cite{xiao2016learning,zhao2017spindle}. We follow the official training and evaluation protocols on Market1501 and CUHK-SYSU, and mainly report the mAP and rank-1 accuracy.
For CUHK03, because we train one single model for all benchmarks, it is slightly different from the standard procedure in \cite{Li2014DeepReID}, which splits the dataset randomly 20 times, and the gallery for testing has 100 identities each time. We only randomly split the dataset once for training and testing, and the gallery includes 200 identities. It means our task might be more difficult than the standard procedure. Similarly, we evaluate our method with rank-1, -5, and -10 accuracy on CUHK03.

\subsection{Implementation Details}
We use Resnet50 and Resnet50-Xception (Resnet-X) pre-trained on ImageNet \cite{russakovsky2015imagenet} as the base models.
Resnet50-Xception replaces the $3 \times 3$ filter kernel through the Xception cell \cite{chollet2016xception}, which contains one $3 \times 3$ channel-wise convolution layer and one $1 \times 1$ spatial convolution layer. Each image is resized into $224\times224$ pixels. The data augmentation includes random horizontal flipping and cropping.
The margins of TriHard loss for both the global and local distances is set to $0.5$, and the mini-batch size is set to $160$, in which each identity has $4$ images.
Each epoch includes 2000 mini-batches. We use an Adam optimizer with an initial learning rate of $10^{-3}$, and shrink this learning rate by a factor of $0.1$ at 80 and 160 epochs until achieving convergence.

For mutual learning, the weight of classification mutual loss (KL) is set to $0.01$, and the weight of metric mutual loss is set to $0.001$.
The optimizer uses Adam with an initial learning rate of $3\times 10^{-4}$, which is reduced to $10^{-4}$ and $10^{-5}$ at 60 epochs and 120 epochs until convergence is achieved.

Re-ranking is an effective technique for boosting the performance of ReID \cite{zhong2017re}. We follow the method and details in \cite{zhong2017re}. In all of our experiments, we combined metric learning loss with classification (identification) loss. 

\subsection{Advantage of AlignedReID}

\begin{figure}[tb]
\centering
\includegraphics[width=1\linewidth]{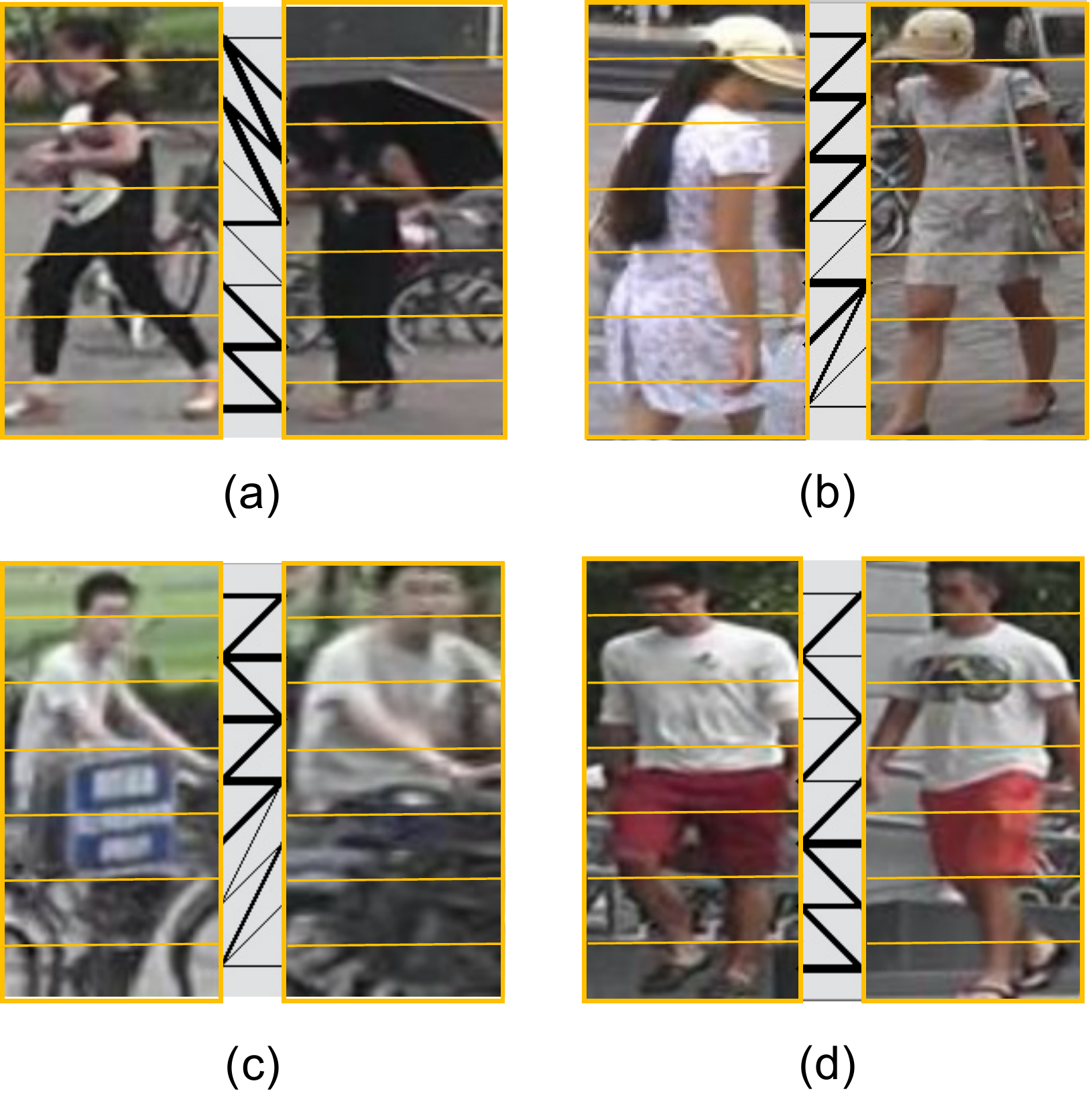}
\caption{The black lines show the alignments of local parts between two persons:
the thicker the line is, the greater it contributes to the shortest path.
Persons have the same identities in (a-c), while persons have different identities in (d).}
\label{fig:SP_result}
\end{figure}

\renewcommand{\multirowsetup}{\centering}  
\begin{table*}[t]\small 
  \begin{center}
  \begin{tabular}{c|c|ccc|ccc|ccc}
    \hline
     							& 			& \multicolumn{3}{c|}{Market1501}	& \multicolumn{3}{c|}{CUHK-SYSU}	& \multicolumn{3}{c}{CUHK03}	\\
    Base model					& Methods 	& mAP 	& r = 1	& r = 5		& mAP 	& r = 1	&r = 5		& r=1 	&r=5		& r = 10	\\
    \hline
    \hline
     \multirow{2}{1.8cm}{Resnet50}	& Baseline	&64.5 	&83.8	&94.1		&88.5	&90.9	&96.6		&83.3	&95.8	&97.9	\\     							
    							& GL-Baseline  &58.0 	&80.4   	&92.0 		&86.0 	&88.2 	&95.6 		&81.7 	&95.0	&97.2	\\
                    					& AlignedReID	&\textbf{72.8} &\textbf{89.2} &\textbf{96.0} &\textbf{92.9} &\textbf{94.5}	&\textbf{98.0}	&\textbf{88.1}	&\textbf{97.5}	&\textbf{98.8}	\\
   \hline
    \multirow{2}{1.8cm}{Resnet50-X} 	& Baseline	&61.7	&83.6	&93.3		&87.9 	&90.4	&95.8	&80.4	&94.5	&97.1	\\      							
    							& GL-Baseline  &57.1 &79.7 &92.4 				&85.9 &87.9 &95.6 &80.7 &94.7 &97.1 \\
                    					& AlignedReID	&\textbf{71.8}& \textbf{89.4}& \textbf{95.8} &\textbf{91.3}	&\textbf{93.5} &\textbf{97.3} &\textbf{88.3} &\textbf{97.1}	&\textbf{98.5}\\  \hline
  \end{tabular}
  \end{center}
  \caption{\label{Table2}Experiment results of AlignedReID. We combine metric learning loss with classification loss in our experiments.}
\end{table*}

\renewcommand{\multirowsetup}{\centering}  
\begin{table*}[t]\small
  \begin{center}
  \begin{tabular}{ c|c|ccc|ccc|ccc}
\hline
    							&	 		& \multicolumn{3}{c|}{Market1501} & \multicolumn{3}{c|}{CUHK-SYSU}	& \multicolumn{3}{c}{CUHK03}	\\
  Loss						&Base model		& mAP 	& r = 1	&r=5 & mAP 	& r = 1	&r=5 & r = 1	&r = 5    &r = 10\\
 	\hline
	\hline
 \multirow{3}{2.3cm}{Baseline} 		&Resnet50	&64.5 	&83.8	&94.1	&88.5	&90.9	&96.6	&83.3	&95.8	&97.9	\\
   										&Resnet50-X	&61.7	&83.6	&93.3	&87.9 	&90.4	&95.8	&80.4	&94.5	&97.1	\\
   \hline
 \multirow{3}{2.3cm}{Baseline+MC} 	&Resnet50	&67.8  	& 86.2	& 94.6  & 89.8 	& 92.2	& 97.1	& 83.8 	& 95.4	& 97.5	\\
   										&Resnet50-X	&68.7 	&87.3	& 95.4	& 89.7 	& 91.8	& 96.8	& 84.6 	& 96.2	& 98.1   \\
   \hline
 \multirow{3}{2.3cm}{Baseline+MM}	&Resnet50	& 66.8 	& 86.3	&95.1	& 90.1 	& 92.2	& 96.9	& 83.8	&95.6 	&97.7\\
 	  						&Resnet50-X	& 66.8 	& 86.2	&94.6	& 90.1 	& 92.5	& 96.8	& 84.2	&95.8 	&97.8\\
 	\hline
	\hline
 \multirow{3}{2.3cm}{AlignedReID} 	&Resnet50	&72.8   &89.2   &96.0   &92.9   &94.5	&98.0	&88.1	&97.5	&98.8	\\
   							&Resnet50-X	&71.8   &89.4   &95.8   &91.3	&93.5   &97.3   &88.3   &97.1	&98.5\\
   \hline
 \multirow{3}{2.3cm}{AlignedReID+MC} &Resnet50	& 70.9 	& 88.6	& 95.9	& 91.4 	& 93.2	& 97.2	& 87.6 	& 97.0	& 98.3	\\
   							&Resnet50-X	& 71.2 	& 88.5	& 95.8	& 91.2 	& 93.3	& 97.4	& 86.9 	& 96.8	& 98.4	\\
   \hline
 \multirow{3}{2.3cm}{AlignedReID+MM}&Resnet50	& 79.3 	& \textbf{91.8}	& \textbf{97.1}	& 94.4 	& 95.7	& 98.8	& \textbf{92.4} 	& 98.9	& 99.5	\\
 	  						&Resnet50-X	& \textbf{79.3} 	& 91.2	& 96.9	& \textbf{94.4} 	& \textbf{95.8}	& \textbf{98.7}	& 92.3 	& \textbf{99.6}	& \textbf{99.8}	\\
   \hline
  \end{tabular}
  \end{center}
  \caption{\label{mutual}Results of mutual learning. MC stands for experiments with classification mutual loss. MM stands for experiments with both classification mutual loss and metric mutual loss.}
  \label{mutual}
\end{table*}

In this section, we analyze the advantage of our AlignedReID model.

We first show some typical results of the alignment in Fig \ref{fig:SP_result}.
In FIg \ref{fig:SP_result}(a), the detection box of the right person is inaccurate, which results in a serious misalignment of heads.
AlignedReID matches the first part of the left image with the first three parts of the right image in the shortest path.
Fig \ref{fig:SP_result}(b) presents another difficult situation where human body structure is defective.
The left image does not contain the parts below the knee.
In the alignment, the skirt side of the right image are associated with the skirt parts of the left one,
while the leg parts of the right image provide small contribution to the shortest path.
Fig \ref{fig:SP_result}(c) shows an example of occlusion, where the lower part of the persons are invisible.
The alignment shows that the occlude parts contribute small in the shortest path, hence the other parts are paid more attention in the learning stage.
Fig \ref{fig:SP_result}(d) show two different persons with similar appearances.
The shirt logo of the right person has no similar part in the left person, which results in a large shortest path distance (local distance) between these two images.

We then compare our \emph{AlignedReID} with two similar networks:
\emph{Baseline} which has no local feature branch, and \emph{GL-Baseline} which has local feature branch without alignment.
In \emph{GL-Baseline}, the local loss is the sum of distances of spatial corresponding local features.
All results are obtained by using the same network and the same training setting. 
The results are shown in Table \ref{Table2}.
Compared to \emph{Baseline}, \emph{GL-Baseline} often gets a worse accuracy. Hence, a local branch without alignment does not help.
Meanwhile, AlignedReID boosts 3.1$\% \sim$ 7.9$\%$ rank-1 accuracy and 3.6$\% \sim$ 10.1$\%$ mAP on all datasets.
The local feature branch with alignment helps the network focus on useful image regions and discriminates similar person images with subtle differences.

We find that if we apply the local distance together with the global distance in the inference stage,
rank-1 accuracy further improves approximately 0.3$\% \sim$ 0.5$\%$. However, it is time consuming and not practical when searching in a large gallery. Hence, we recommend using the global feature only.

\subsection{Analysis of Mutual Learning}
In the mutual learning experiment, we simultaneously train two AlignedReID models.
One model is based on Resnet50, and the other is based on Resnet50-Xception.
We compare their performances for three cases: with both metric mutual loss and classification mutual loss, with only classification mutual loss, and with no mutual loss.
We also conduct a similar mutual learning experiment as a baseline, where the global features are trained without local features. The results are shown in Table \ref{mutual}.

Both experiments show that the metric mutual learning method can further improve performance.
With the baseline mutual learning experiment, the classification mutual loss significantly improves performance on all datasets.
However, with the AlignedReID mutual learning experiment, because the models without mutual learning perform well enough, the classification mutual loss cannot further improve performance.

\subsection{Comparison with Other Methods}
In this subsection, we compare the results of AlignedReID with state-of-the-art methods, in Table \ref{table_market1501} $\sim$ \ref{table_cuhksysu}.
In the tables, AlignedReID represents our method with mutual learning, and AlignedReID (RK) is our method with both mutual learning and re-ranking \cite{zhong2017re} with $k$-reciprocal encoding.

On Market1501, GLAD \cite{wei2017glad} achieves an 89.9$\%$ rank-1 accuracy, and our AlignedReID achieves a 91.8$\%$ rank-1 accuracy, exceeding it.
For mAP, \cite{hermans2017defense} obtains 81.1$\%$ owing to the use of re-ranking.
With the help of re-ranking, the rank-1 accuracy and mAP are further improved to 94.4$\%$ and 90.7$\%$ in our AlignedReID (RK),
outperforming the best of previous works by 4.5$\%$ and 9.6$\%$ separately.

On CUHK03, without re-ranking, HydraPlus-Net \cite{Liu2017HydraPlus} achieves 91.8$\%$ rank-1 accuracy and our AlignedReID yields 92.4$\%$.
Note that our test gallery size is two times large as that used in \cite{Liu2017HydraPlus}.
Furthermore, our AlignedReID (RK) obtains a 97.8$\%$ rank-1 accuracy, exceeding state-of-the-art by 6.0$\%$.

There have not been many studies reported on CUHK-SYSU.
With this dataset, AlignedReID achieves 94.4$\%$ mAP and 95.8$\%$ rank-1 accuracy, which is much higher than any published results.

\begin{table}[htb]
  \caption{\label{table_market1501}Comparison on \textbf{Market1501} in single query mode}
  \begin{center}
  \begin{tabular}{c|cc}
    	\hline
	Methods  							& mAP	& r=1			\\
	\hline
	\hline
	Temporal \cite{martinel2016temporal}	&22.3	&47.9	\\
	Learning \cite{zhang2016learning}		&35.7	&61.0	\\
	Gated \cite{varior2016gated}			&39.6	&65.9		\\
	Person \cite{chen2017person}			&45.5	&71.8		\\
	Re-ranking \cite{zhong2017re}			&63.6	&77.1	\\
	Pose \cite{zheng2017pose}			&56.0	&79.3		\\
	Scalable \cite{bai2017scalable}			&68.8	&82.2	\\
	Improving \cite{lin2017improving}		&64.7	&84.3	\\
	In \cite{hermans2017defense}			&69.1	&84.9		\\
	In (RK)\cite{hermans2017defense}		&\textbf{81.1}	&86.7	\\
	Spindle\cite{zhao2017spindle}			&-		&76.9	\\
	Deep\cite{zhang2017deep}$^*$		&68.8	&87.7	\\
	DarkRank\cite{chen2017darkrank}$^*$	&74.3	&89.8	\\
	GLAD\cite{wei2017glad}$^*$          &73.9   &\textbf{89.9} \\
    HydraPlus-Net\cite{Liu2017HydraPlus}$^*$	&-		&76.9	\\
   	\hline
	AlignedReID							& 79.3 	& 91.8	\\
	AlignedReID (RK)						&\color{red}{\textbf{90.7}}	&\color{red}{\textbf{94.4}}\\
  \hline
  \end{tabular}
  \end{center}
  \label{market1501}
\end{table}

\begin{table}[htb]
  \caption{\label{table_cuhk03}Comparison on \textbf{CUHK03} labeled dataset}
  \begin{center}
  \begin{tabular}{c|ccc}
    	\hline
	Methods  							& r=1	& r=5	&r=10		\\
	\hline
	\hline
	Person \cite{liao2015person} 			& 44.6	& -		&-\\
	Learning \cite{zhang2016learning}		& 62.6	&90.0 	&94.8 \\
	Gated \cite{varior2016gated} 			& 61.8	& -		&-\\
	A \cite{varior2016a} 					& 57.3  	& 80.1	&88.3\\
	Re-ranking \cite{zhong2017re}			& 64.0	& -		&-\\
	In \cite{hermans2017defense} 			& 75.5	&95.2	&99.2\\
	Joint \cite{xiao2017joint}				& 77.5	& -		&-\\
	Deep \cite{geng2016deep}$^*$			& 84.1	& -		&-\\
	Looking \cite{barbosa2017looking}$^*$	& 72.4	&95.2	&95.8\\
	Unlabeled \cite{Zheng_2017_ICCV}	& 84.6 	& 97.6	&98.9\\
	A \cite{zheng2016discriminatively}$^*$	& 83.4	& 97.1	&98.7\\
	Spindle\cite{zhao2017spindle}			& 88.5	&97.8	&98.6\\
	DarkRank\cite{chen2017darkrank}$^*$	&89.7	&\textbf{98.4}	&\textbf{99.2}\\
	GLAD\cite{wei2017glad}$^*$          &85.0   &97.9      &99.1\\
    HydraPlus-Net\cite{Liu2017HydraPlus}$^*$	&\textbf{91.8}	&\textbf{98.4}	&99.1	\\
	\hline
	AlignedReID							& 92.4	&98.9 	&99.5	\\
	AlignedReID (RK)						&\color{red}{\textbf{97.8}}	&\color{red}{\textbf{99.6}} &\color{red}{\textbf{99.8}}\\
  \hline
  \end{tabular}
  \end{center}
  \label{cuhk03}
\end{table}


\begin{table}[htb]
  \caption{\label{table_cuhksysu}Comparison with existing methods on \textbf{CUHK-SYSU}}
  \begin{center}
  \begin{tabular}{c|cc}
    	\hline
	Methods  						& mAP	& r=1			\\
	\hline
	\hline
	End\cite{xiao2016end}			& 55.7&62.7		\\
	Deep \cite{schumann2016deep}$^*$&74.0&76.7 \\
	Neural \cite{Liu_2017_ICCV} 		&\textbf{77.9} &\textbf{81.2}\\
   	\hline
	AlignedReID						&\color{red}{\textbf{94.4}}&\color{red}{\textbf{95.7}} \\
  \hline
  \end{tabular}
  \end{center}
  \label{cuhk-sysu}
\end{table}

\begin{figure*}[htb]
\centering
\includegraphics[width=0.99\linewidth]{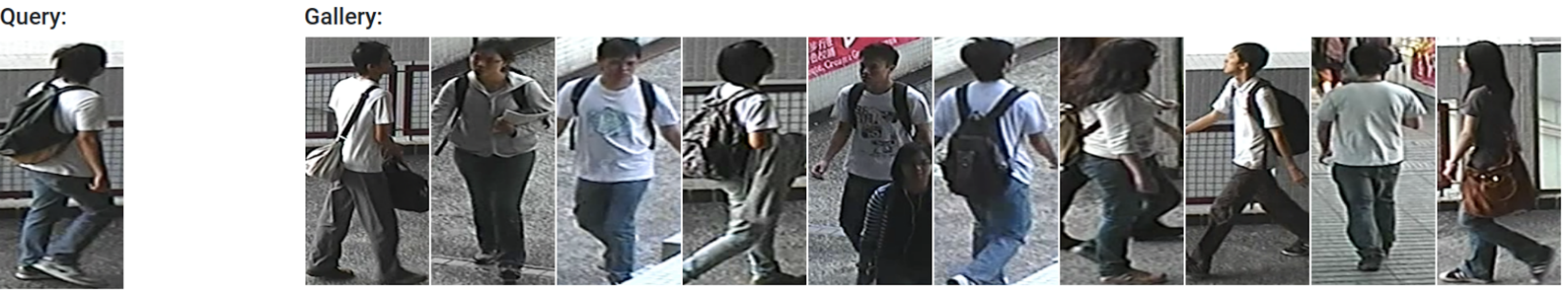}
\caption{Interface of our human performance evaluation system for CUHK03. The left side shows a query image and the right side shows 10 images sampled using our deep model.}
\label{fig:HP}
\end{figure*}

\section{Human Performance in Person ReID}
\label{humanperformance}
Given the significant improvement of our approach, we are curious to
find the quality of human performance.
Thus, we conduct human performance evaluations on Market1501 and CUHK03.

To make the study feasible, for each query image, the annotator does not have to find the same person from the entire gallery set. We ask him or her to pick the answer from a much smaller set of selected images.

In CUHK03, for each query image, there is only one image for the identical person in the gallery set.
The annotator looks for the identical person among $10$ images selected:
our ReID model first generates the top10 results in the gallery set for the query image;
if the ``ground truth" is not among the top10 results, we replace the $10$th result with the ground truth.

For Market1501, there may be more than one ground truth in the gallery set.
The annotator needs to pick one from $50$ images selected as follows:
our ReID model generated the top50 results in the gallery set for the query image;
if any ground truth is not among them, it would be used to replace one non-ground truth result with the lowest rank.
In this way, we make sure that all ground truths are in the $50$ selected images.

The interface of the human performance evaluation system is presented in Fig \ref{fig:HP}.
The images are randomly shuffled before being displayed to the annotator. The evaluation website will be available soon.
Ten professional annotators participate in the evaluation. Because only one candidate is chosen, we are unable to obtain the mAP of human beings as a standard evaluation.
The rank-1 accuracies are computed for each annotator on all datasets.
The best accuracy is then used as the human performance, which is shown in Table \ref{Table:HP}.

On Market1501, human beings achieve a 93.5$\%$ rank-1 accuracy, which is better than all state-of-the-art methods.
The rank-1 accuracy in our AlignedReID (RK) reaches 94.4$\%$ rank-1, exceeding the human performance.
On CUHK03, the human performance reaches a 95.7$\%$ rank-1 accuracy, which is much higher than any known state-of-the-art methods.
Our AlignedReID (RK) obtains a 97.8$\%$ rank-1 accuracy, surpassing the human performance.

\begin{figure}[tbh]
\centering
\includegraphics[width=0.9\linewidth]{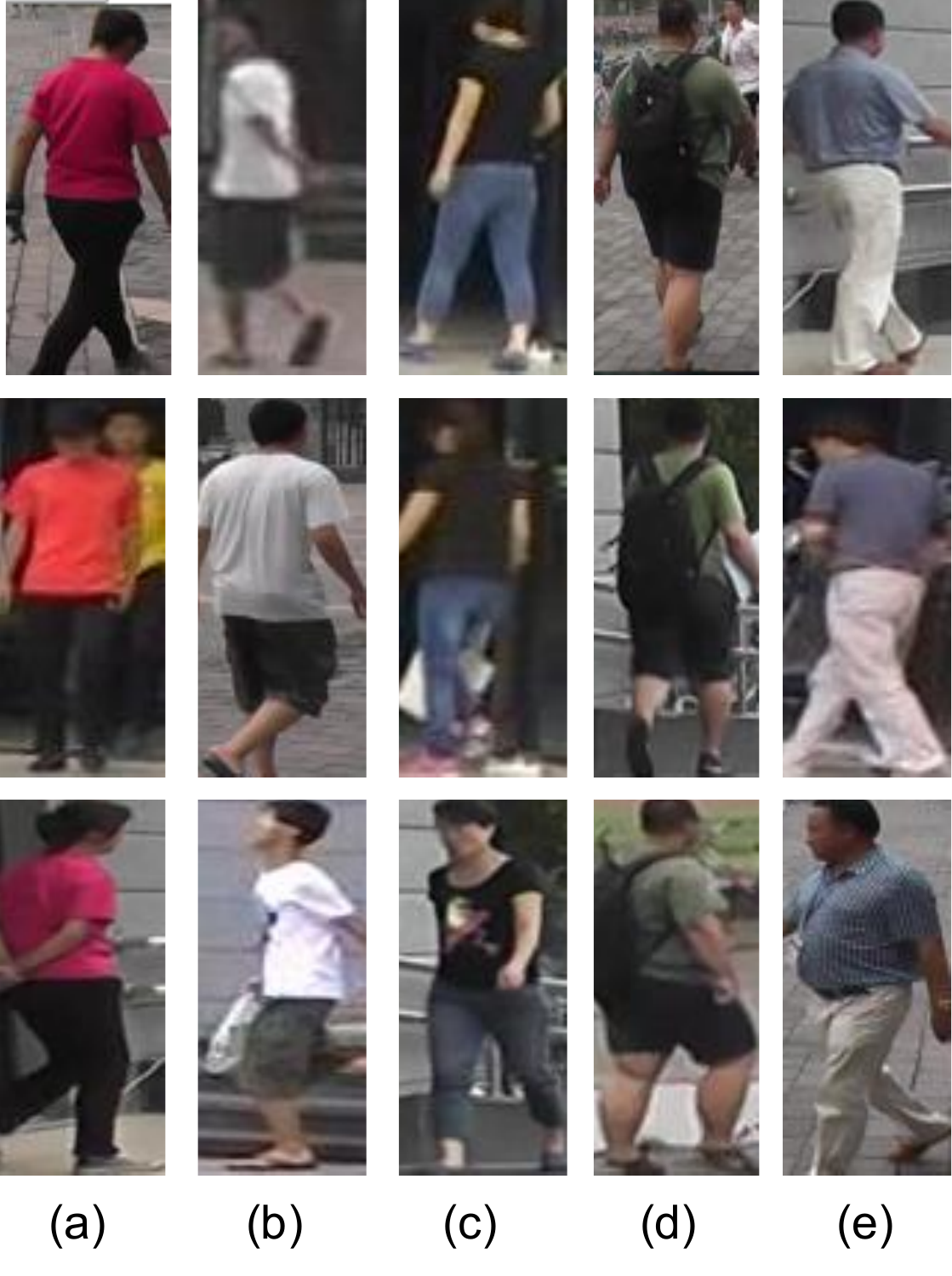}
\caption{Top: query image. Middle: the result picked by an annotator. Bottom: top1 result by our method.}
\label{fig:out_human_example}
\end{figure}

Figure \ref{fig:out_human_example} shows some examples, where an annotator selected a wrong answer, while the top1 result provided by our method is correct.

\begin{table}[htb]
  \caption{\label{Table:HP}Results of human performance evaluation.
          We show the accuracies of the five annotators who did best in the evaluation.
          We also show our AlignedReID results with re-ranking.}
   \begin{center}
   \begin{tabular}{c |c |c  }
  	\hline
			&Market1501 &CUHK03 \\
	\hline
    \hline
    Human Rank 1		    &\textbf{93.5}       &\textbf{95.7}	\\
    Human Rank 2		    &91.1       &91.9    \\
    Human Rank 3    		&90.6       &91.2    \\
    Human Rank 4    		&90.0       &91.1    \\
    Human Rank 5    		&88.3       &90.0    \\
    \hline
	AlignedReID (RK) &\color{red}{\textbf{94.4}}  &\color{red}{\textbf{97.8}} \\
    \hline
    \end{tabular}
    \end{center}
\end{table}

%

\section{Conclusion}
\label{conclusion}

In this paper, we have demonstrated that an implicit alignment of local features can substantially improve global feature learning. This surprising result gives us an important insight: the end-to-end learning with structure prior is more powerful than a ``blind" end-to-end learning. 

\begin{figure}[tbh]
\centering
\includegraphics[width=0.9\linewidth]{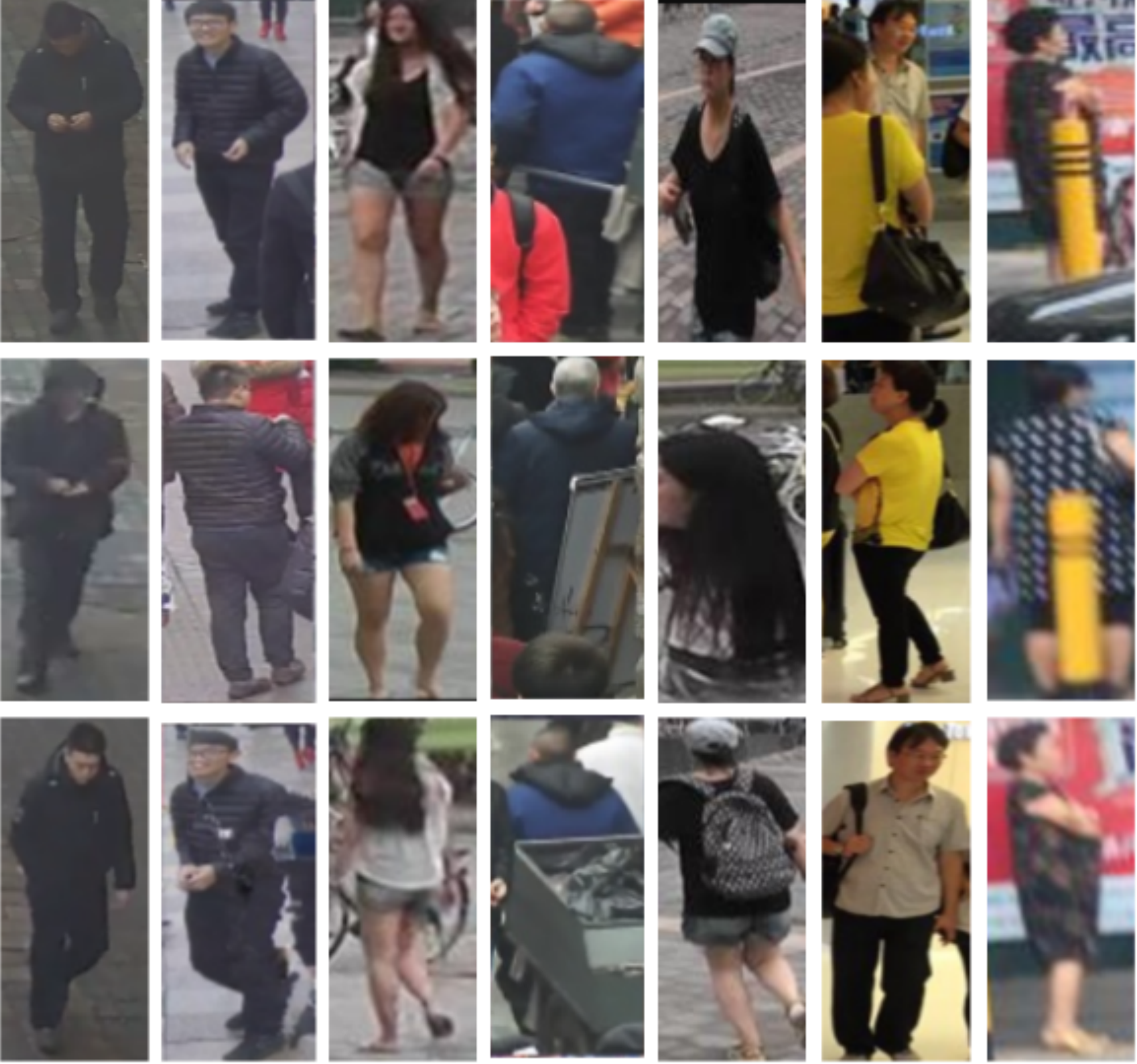}
\caption{Top: query image. Middle: top1 result by our method. Bottom: ground truth.}
\label{fig:error}
\end{figure}

Although we show that our methods outperform humans in the Market1501 and CUHK03 datasets,
it is still early to claim that machines beat humans in general.
Figure \ref{fig:error} presents a few ``big" mistakes which seldom confuses humans.
This indicates that the machine still has a lot of room for improvement.

%

{\small
\bibliographystyle{ieee}
\bibliography{ref}

\begin{thebibliography}{10}\itemsep=-1pt

\bibitem{bai2017scalable}
S.~Bai, X.~Bai, and Q.~Tian.
\newblock Scalable person re-identification on supervised smoothed manifold.
\newblock {\em arXiv preprint arXiv:1703.08359}, 2017.

\bibitem{barbosa2017looking}
I.~B. Barbosa, M.~Cristani, B.~Caputo, A.~Rognhaugen, and T.~Theoharis.
\newblock Looking beyond appearances: Synthetic training data for deep cnns in
  re-identification.
\newblock {\em arXiv preprint arXiv:1701.03153}, 2017.

\bibitem{chen2017beyond}
W.~Chen, X.~Chen, J.~Zhang, and K.~Huang.
\newblock Beyond triplet loss: a deep quadruplet network for person
  re-identification.
\newblock {\em arXiv preprint arXiv:1704.01719}, 2017.

\bibitem{chen2017darkrank}
Y.~Chen, N.~Wang, and Z.~Zhang.
\newblock Darkrank: Accelerating deep metric learning via cross sample
  similarities transfer.
\newblock {\em arXiv preprint arXiv:1707.01220}, 2017.

\bibitem{chen2017person}
Y.-C. Chen, X.~Zhu, W.-S. Zheng, and J.-H. Lai.
\newblock Person re-identification by camera correlation aware feature
  augmentation.
\newblock {\em IEEE Transactions on Pattern Analysis and Machine Intelligence},
  2017.

\bibitem{cheng2016person}
D.~Cheng, Y.~Gong, S.~Zhou, J.~Wang, and N.~Zheng.
\newblock Person re-identification by multi-channel parts-based cnn with
  improved triplet loss function.
\newblock In {\em Proceedings of the IEEE Conference on Computer Vision and
  Pattern Recognition}, pages 1335--1344, 2016.

\bibitem{chollet2016xception}
F.~Chollet.
\newblock Xception: Deep learning with depthwise separable convolutions.
\newblock {\em arXiv preprint arXiv:1610.02357}, 2016.

\bibitem{fan2017unsupervised}
H.~Fan, L.~Zheng, and Y.~Yang.
\newblock Unsupervised person re-identification: Clustering and fine-tuning.
\newblock {\em arXiv preprint arXiv:1705.10444}, 2017.

\bibitem{farenzena2010person}
M.~Farenzena, L.~Bazzani, A.~Perina, V.~Murino, and M.~Cristani.
\newblock Person re-identification by symmetry-driven accumulation of local
  features.
\newblock In {\em Computer Vision and Pattern Recognition (CVPR), 2010 IEEE
  Conference on}, pages 2360--2367. IEEE, 2010.

\bibitem{geng2016deep}
M.~Geng, Y.~Wang, T.~Xiang, and Y.~Tian.
\newblock Deep transfer learning for person re-identification.
\newblock {\em arXiv preprint arXiv:1611.05244}, 2016.

\bibitem{hamdoun2008person}
O.~Hamdoun, F.~Moutarde, B.~Stanciulescu, and B.~Steux.
\newblock Person re-identification in multi-camera system by signature based on
  interest point descriptors collected on short video sequences.
\newblock In {\em Distributed Smart Cameras, 2008. ICDSC 2008. Second ACM/IEEE
  International Conference on}, pages 1--6. IEEE, 2008.

\bibitem{he2016deep}
K.~He, X.~Zhang, S.~Ren, and J.~Sun.
\newblock Deep residual learning for image recognition.
\newblock In {\em Proceedings of the IEEE conference on computer vision and
  pattern recognition}, pages 770--778, 2016.

\bibitem{hermans2017defense}
A.~Hermans, L.~Beyer, and B.~Leibe.
\newblock In defense of the triplet loss for person re-identification.
\newblock {\em arXiv preprint arXiv:1703.07737}, 2017.

\bibitem{Li2014DeepReID}
W.~Li, R.~Zhao, T.~Xiao, and X.~Wang.
\newblock Deepreid: Deep filter pairing neural network for person
  re-identification.
\newblock pages 152--159, 2014.

\bibitem{liao2015person}
S.~Liao, Y.~Hu, X.~Zhu, and S.~Z. Li.
\newblock Person re-identification by local maximal occurrence representation
  and metric learning.
\newblock In {\em Proceedings of the IEEE Conference on Computer Vision and
  Pattern Recognition}, pages 2197--2206, 2015.

\bibitem{lin2017improving}
Y.~Lin, L.~Zheng, Z.~Zheng, Y.~Wu, and Y.~Yang.
\newblock Improving person re-identification by attribute and identity
  learning.
\newblock {\em arXiv preprint arXiv:1703.07220}, 2017.

\bibitem{Liu_2017_ICCV}
H.~Liu, J.~Feng, Z.~Jie, K.~Jayashree, B.~Zhao, M.~Qi, J.~Jiang, and S.~Yan.
\newblock Neural person search machines.
\newblock In {\em The IEEE International Conference on Computer Vision (ICCV)},
  Oct 2017.

\bibitem{liu2017end}
H.~Liu, J.~Feng, M.~Qi, J.~Jiang, and S.~Yan.
\newblock End-to-end comparative attention networks for person
  re-identification.
\newblock {\em IEEE Transactions on Image Processing}, 2017.

\bibitem{liu2017video}
H.~Liu, Z.~Jie, K.~Jayashree, M.~Qi, J.~Jiang, S.~Yan, and J.~Feng.
\newblock Video-based person re-identification with accumulative motion
  context.
\newblock {\em arXiv preprint arXiv:1701.00193}, 2017.

\bibitem{Liu2017HydraPlus}
X.~Liu, H.~Zhao, M.~Tian, L.~Sheng, J.~Shao, S.~Yi, J.~Yan, and X.~Wang.
\newblock Hydraplus-net: Attentive deep features for pedestrian analysis.
\newblock 2017.

\bibitem{liu2017quality}
Y.~Liu, J.~Yan, and W.~Ouyang.
\newblock Quality aware network for set to set recognition.
\newblock {\em arXiv preprint arXiv:1704.03373}, 2017.

\bibitem{ma2017person}
X.~Ma, X.~Zhu, S.~Gong, X.~Xie, J.~Hu, K.-M. Lam, and Y.~Zhong.
\newblock Person re-identification by unsupervised video matching.
\newblock {\em Pattern Recognition}, 65:197--210, 2017.

\bibitem{martinel2016temporal}
N.~Martinel, A.~Das, C.~Micheloni, and A.~K. Roy-Chowdhury.
\newblock Temporal model adaptation for person re-identification.
\newblock In {\em European Conference on Computer Vision}, pages 858--877.
  Springer, 2016.

\bibitem{matsukawa2016person}
T.~Matsukawa and E.~Suzuki.
\newblock Person re-identification using cnn features learned from combination
  of attributes.
\newblock In {\em Pattern Recognition (ICPR), 2016 23rd International
  Conference on}, pages 2428--2433. IEEE, 2016.

\bibitem{mclaughlin2016recurrent}
N.~McLaughlin, J.~Martinez~del Rincon, and P.~Miller.
\newblock Recurrent convolutional network for video-based person
  re-identification.
\newblock In {\em Proceedings of the IEEE Conference on Computer Vision and
  Pattern Recognition}, pages 1325--1334, 2016.

\bibitem{peng2016unsupervised}
P.~Peng, T.~Xiang, Y.~Wang, M.~Pontil, S.~Gong, T.~Huang, and Y.~Tian.
\newblock Unsupervised cross-dataset transfer learning for person
  re-identification.
\newblock In {\em Proceedings of the IEEE Conference on Computer Vision and
  Pattern Recognition}, pages 1306--1315, 2016.

\bibitem{radenovic2016cnn}
F.~Radenovi{\'c}, G.~Tolias, and O.~Chum.
\newblock Cnn image retrieval learns from bow: Unsupervised fine-tuning with
  hard examples.
\newblock In {\em European Conference on Computer Vision}, pages 3--20.
  Springer, 2016.

\bibitem{russakovsky2015imagenet}
O.~Russakovsky, J.~Deng, H.~Su, J.~Krause, S.~Satheesh, S.~Ma, Z.~Huang,
  A.~Karpathy, A.~Khosla, M.~Bernstein, et~al.
\newblock Imagenet large scale visual recognition challenge.
\newblock {\em International Journal of Computer Vision}, 115(3):211--252,
  2015.

\bibitem{schumann2016deep}
A.~Schumann, S.~Gong, and T.~Schuchert.
\newblock Deep learning prototype domains for person re-identification.
\newblock {\em arXiv preprint arXiv:1610.05047}, 2016.

\bibitem{zheng2016mars}
Springer.
\newblock {\em MARS: A Video Benchmark for Large-Scale Person
  Re-identification}, 2016.

\bibitem{tesfaye2017multi}
Y.~T. Tesfaye, E.~Zemene, A.~Prati, M.~Pelillo, and M.~Shah.
\newblock Multi-target tracking in multiple non-overlapping cameras using
  constrained dominant sets.
\newblock {\em arXiv preprint arXiv:1706.06196}, 2017.

\bibitem{varior2016gated}
R.~R. Varior, M.~Haloi, and G.~Wang.
\newblock Gated siamese convolutional neural network architecture for human
  re-identification.
\newblock In {\em European Conference on Computer Vision}, pages 791--808.
  Springer, 2016.

\bibitem{varior2016siamese}
R.~R. Varior, B.~Shuai, J.~Lu, D.~Xu, and G.~Wang.
\newblock A siamese long short-term memory architecture for human
  re-identification.
\newblock In {\em European Conference on Computer Vision}, pages 135--153.
  Springer, 2016.

\bibitem{varior2016a}
R.~R. Varior, B.~Shuai, J.~Lu, D.~Xu, and G.~Wang.
\newblock A siamese long short-term memory architecture for human
  re-identification.
\newblock In {\em European Conference on Computer Vision}, pages 135--153,
  2016.

\bibitem{wang2017deep}
J.~Wang, S.~Zhou, J.~Wang, and Q.~Hou.
\newblock Deep ranking model by large adaptive margin learning for person
  re-identification.
\newblock {\em arXiv preprint arXiv:1707.00409}, 2017.

\bibitem{wang2016person}
T.~Wang, S.~Gong, X.~Zhu, and S.~Wang.
\newblock Person re-identification by discriminative selection in video
  ranking.
\newblock {\em IEEE transactions on pattern analysis and machine intelligence},
  38(12):2501--2514, 2016.

\bibitem{wei2017glad}
L.~Wei, S.~Zhang, H.~Yao, W.~Gao, and Q.~Tian.
\newblock Glad: Global-local-alignment descriptor for pedestrian retrieval.
\newblock {\em arXiv preprint arXiv:1709.04329}, 2017.

\bibitem{xiao2016cross}
Q.~Xiao, K.~Cao, H.~Chen, F.~Peng, and C.~Zhang.
\newblock Cross domain knowledge transfer for person re-identification.
\newblock {\em arXiv preprint arXiv:1611.06026}, 2016.

\bibitem{xiao2017margin}
Q.~Xiao, H.~Luo, and C.~Zhang.
\newblock Margin sample mining loss: A deep learning based method for person
  re-identification.
\newblock {\em arXiv preprint arXiv:1710.00478}, 2017.

\bibitem{xiao2016learning}
T.~Xiao, H.~Li, W.~Ouyang, and X.~Wang.
\newblock Learning deep feature representations with domain guided dropout for
  person re-identification.
\newblock In {\em Proceedings of the IEEE Conference on Computer Vision and
  Pattern Recognition}, pages 1249--1258, 2016.

\bibitem{xiao2016end}
T.~Xiao, S.~Li, B.~Wang, L.~Lin, and X.~Wang.
\newblock End-to-end deep learning for person search.
\newblock {\em arXiv preprint arXiv:1604.01850}, 2016.

\bibitem{xiao2017joint}
T.~Xiao, S.~Li, B.~Wang, L.~Lin, and X.~Wang.
\newblock Joint detection and identification feature learning for person
  search.
\newblock In {\em Proc. CVPR}, 2017.

\bibitem{yao2017deep}
H.~Yao, S.~Zhang, Y.~Zhang, J.~Li, and Q.~Tian.
\newblock Deep representation learning with part loss for person
  re-identification.
\newblock {\em arXiv preprint arXiv:1707.00798}, 2017.

\bibitem{you2016top}
J.~You, A.~Wu, X.~Li, and W.-S. Zheng.
\newblock Top-push video-based person re-identification.
\newblock In {\em Proceedings of the IEEE Conference on Computer Vision and
  Pattern Recognition}, pages 1345--1353, 2016.

\bibitem{zhang2017image}
D.~Zhang, W.~Wu, H.~Cheng, R.~Zhang, Z.~Dong, and Z.~Cai.
\newblock Image-to-video person re-identification with temporally memorized
  similarity learning.
\newblock {\em IEEE Transactions on Circuits and Systems for Video Technology},
  2017.

\bibitem{Zhang_2016_CVPR}
L.~Zhang, T.~Xiang, and S.~Gong.
\newblock Learning a discriminative null space for person re-identification.
\newblock In {\em The IEEE Conference on Computer Vision and Pattern
  Recognition (CVPR)}, June 2016.

\bibitem{zhang2016learning}
L.~Zhang, T.~Xiang, and S.~Gong.
\newblock Learning a discriminative null space for person re-identification.
\newblock In {\em Proceedings of the IEEE Conference on Computer Vision and
  Pattern Recognition}, pages 1239--1248, 2016.

\bibitem{zhang2017learning}
W.~Zhang, S.~Hu, and K.~Liu.
\newblock Learning compact appearance representation for video-based person
  re-identification.
\newblock {\em arXiv preprint arXiv:1702.06294}, 2017.

\bibitem{zhang2017deep}
Y.~Zhang, T.~Xiang, T.~M. Hospedales, and H.~Lu.
\newblock Deep mutual learning.
\newblock {\em arXiv preprint arXiv:1706.00384}, 2017.

\bibitem{zhao2017spindle}
H.~Zhao, M.~Tian, S.~Sun, J.~Shao, J.~Yan, S.~Yi, X.~Wang, and X.~Tang.
\newblock Spindle net: Person re-identification with human body region guided
  feature decomposition and fusion.
\newblock CVPR, 2017.

\bibitem{zhao2017person}
R.~Zhao, W.~Oyang, and X.~Wang.
\newblock Person re-identification by saliency learning.
\newblock {\em IEEE transactions on pattern analysis and machine intelligence},
  39(2):356--370, 2017.

\bibitem{zheng2017pose}
L.~Zheng, Y.~Huang, H.~Lu, and Y.~Yang.
\newblock Pose invariant embedding for deep person re-identification.
\newblock {\em arXiv preprint arXiv:1701.07732}, 2017.

\bibitem{zheng2015scalable}
L.~Zheng, L.~Shen, L.~Tian, S.~Wang, J.~Wang, and Q.~Tian.
\newblock Scalable person re-identification: A benchmark.
\newblock In {\em Computer Vision, IEEE International Conference}, 2015.

\bibitem{zheng2016person}
L.~Zheng, Y.~Yang, and A.~G. Hauptmann.
\newblock Person re-identification: Past, present and future.
\newblock {\em arXiv preprint arXiv:1610.02984}, 2016.

\bibitem{zheng2016discriminatively}
Z.~Zheng, L.~Zheng, and Y.~Yang.
\newblock A discriminatively learned cnn embedding for person
  re-identification.
\newblock {\em arXiv preprint arXiv:1611.05666}, 2016.

\bibitem{Zheng_2017_ICCV}
Z.~Zheng, L.~Zheng, and Y.~Yang.
\newblock Unlabeled samples generated by gan improve the person
  re-identification baseline in vitro.
\newblock In {\em The IEEE International Conference on Computer Vision (ICCV)},
  Oct 2017.

\bibitem{zhong2017re}
Z.~Zhong, L.~Zheng, D.~Cao, and S.~Li.
\newblock Re-ranking person re-identification with k-reciprocal encoding.
\newblock {\em arXiv preprint arXiv:1701.08398}, 2017.

\bibitem{zhousee}
Z.~Zhou, Y.~Huang, W.~Wang, L.~Wang, and T.~Tan.
\newblock See the forest for the trees: Joint spatial and temporal recurrent
  neural networks for video-based person re-identification.

\end{thebibliography}
}

\end{document}